*Research Paper*

# Reshaping Smart Energy Transition:

## An analysis of human-building interactions in Qatar Using Machine Learning Techniques

Rateb Jabbar, College of Arts and Science, Qatar University, Doha, Qatar  
Esmat Zaidan, College of Arts and Science, Qatar University, Doha, Qatar  
Ahmed ben Said, College of Engineering, Qatar University, Doha, Qatar  
Ali Ghofrani, College of Arts and Science, Qatar University, Doha, Qatar

**Abstract**

*Policy Planning have the potential to contribute to the strategic development and economic diversification of developing countries even without considerable structural changes. In this study, we analyzed a set of human-oriented dimensions aimed at improving energy policies related to the building sector in Qatar. Considering the high percentage of expatriate and migrant communities with different financial and cultural backgrounds and behavioral patterns compared with local communities in the GCC Union, it is required to investigate human dimensions to propose adequate energy policies. This study explored the correlations of socioeconomic, behavioral, and demographic dimensions to determine the main factors behind discrepancies in energy use, responsibilities, motivations, habits, and overall well-being. The sample included 2,200 people in Qatar, and it was clustered into two consumer categories: high and low. In particular, the study focused on exploring human indoor comfort perception dependencies with building features. Financial drivers, such as demand programs and energy subsidies, were explored in relation to behavioral patterns. Subsequently, the data analysis resulted in implications for energy policies regarding interventions, social well-being, and awareness. Machine learning methods were used to perform a feature importance analysis to determine the main factors of human behavior. The findings of this study demonstrated how human factors impact comfort perception in residential and work environments, norms, habits, self-responsibility, consequence awareness, and consumption. The study has important implications for developing targeted strategies aimed at improving the efficacy of energy policies and sustainability performance indicators.*

## 1. Introduction

The development of a knowledge-based economy represents the core of Qatar's strategies, with a particular focus on social well-being and sustainability (unescwa, 2017). Economic growth rates in Qatar have been among the fastest globally in the last two decades, which considerably impacted its energy demands. The population growth as well has contributed to the increasing number of installed electricity meters in the commercial and residential sectors and electricity consumption. Moreover, the arid climate presents a substantial cooling load in the building sector. To illustrate, 46% to 61% of the annual electricity is consumed for cooling purposes, as the building simulations of eight residential and commercial buildings demonstrated (Nazemi, Zaidan and Jafari, 2021). Up to date, the strategies have not included human-building interactions (HBIs) (Labanca and Bertoldi, 2018), although residents indirectly and directly interact with equipment and buildings (Angizeh et al., 2021)(Zaidan and Abulibdeh, 2021). Accordingly, performance gaps are a result of non-including human factors in energy decision-making (Bertoldi, 2019)(Jafari et al., 2020)(Zaidan, Al-Saidi and Hammad, 2019). HBIs are critical for the building energy efficiency and conservation equation (B and Lalanne, 2017). Relevant literature has identified human factors that impact energy use (Jia, Srinivasan and Raheem, 2017),(Zhao and Magoulès, 2012). To illustrate, occupants with 'wasteful workstyles' consume up to double the energy that those with "non-wasteful" styles (Lin and Hong, 2013). Moreover, studies have demonstrated that occupants' preferences depend on gender, age, and cultural and psychological factors (Masoso and Grobler, 2010), (Thiaux *et al.*, 2019) this and can predict the operation of commercial buildings (Nematchoua et al., 2019),(Appel-Meulenbroek and Danivska, 2021),(Nazemi, Zaidan and Jafari, 2021) such as offices. According to some researchers, human factors in commercial buildings are the 'dark side' of energy use (Masoso and Grobler, 2010). Accordingly, the implications of occupant behavior are stochastic and require in-depth investigation (Hong et al., 2017),(Thiaux et al., 2019), because technology investments are not sufficient to ensure higher comfort perceptions and low or net-zero energy in the buildings (D'Oca, Hong and Langevin, 2018).

Our study surveyed 2,200 individuals in Doha, Qatar to determine the interdependencies in HBIs using empirical and analytical analysis. The sample was clustered using the k-means clustering method into two consumer categories in terms of the energy use intensity of their dwellings. We conducted a feature importance analysis based on a random forest classification to determine the main factors behind human energy consumption patterns. Subsequently, we classified the sample data on the basis of the primary human-oriented factors to identify the relationships between the survey elements. The results were then used to propose recommendations to integrate human dimensions into energy policy and improve the outcomes of strategic plans for the building environment sector in Qatar.

## 2. Methodology and Survey Structure

This study explored human and social factors in improving the efficiency of energy policy and overcoming performance gaps in the building sector in Qatar. Accordingly, we surveyed the sample in terms of demographic (gender, age, and ethnicity) and socioeconomic (employment status, marital status, expenses, and household income) factors and their association with behavioral factors and human interactions in building environments. We used both quantitative and qualitative methods to identify indoor lighting and thermal comfort and indoor environment preferences and to determine how they impact respondents' performance and well-being. We investigated the habits and routines of the respondents in interacting with the building, such as thermostat switching and window use, to determine relationships between features. In addition, we explored behavioral and psychological factors, including attitudes, motivations, responsibilities, awareness of consequences, and their intercorrelations. The main factors are shown in **Error! Reference source not found.**, as well as the human-oriented applications based on particular demographic and socioeconomic characteristics of the migrant and local communities.

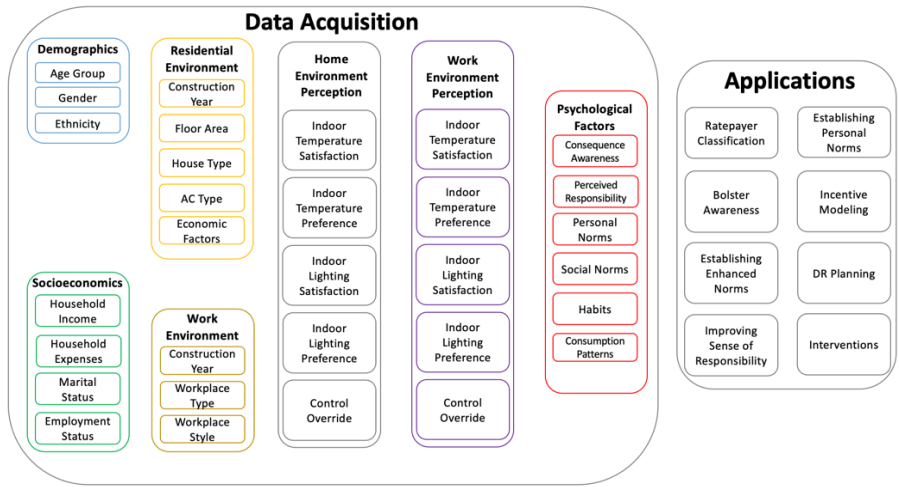

Figure 1: Survey framework comprising the data acquisition of human dimensions and building properties of the sample

## 3. Categorization of Energy Consumers

Determining primary human-driven factors that impact consumption habits and patterns is required for improving energy policies by targeting specific population segments and developing strategies through coercive actions, pricing, interventions, incentive modeling, and awareness raising. This is critical for Qatar considering its variety in human characteristics.

### 3.1. Consumer clustering

To compare consumption patterns and calculate a new attribute equivalent to energy use intensity (EUI), we divided the average monthly electricity payments by floor area to calculate a new attribute equivalent to energy use intensity (EUI). Qatar's residential rate structure is not based on dynamic pricing schemes. Subsequently, the k-means clustering method (Jin and Han, 2010) was used to categorize EUI values into low and high consumers. We partitioned 1021 respondents into 829 users of Category 1 ($EUI$=2.89$QRm$2) representing low consumers and 192 users in Category 2, representing high consumers. Two patterns are illustrated in **Error! Reference source not found.** and **Error! Reference source not found.**. Subsequently, we used those patterns to perform a feature importance analysis.

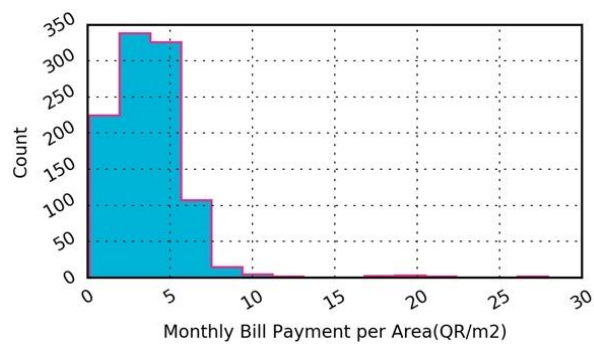

Figure 2: Distribution of users' energy use intensity on the basis of average monthly bill payments and home floor area.

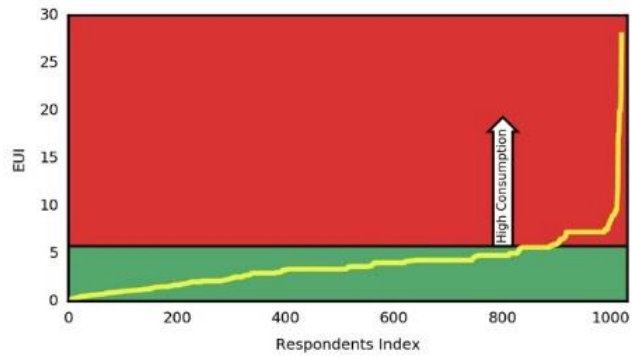

**Figure 3:** Low (green area) and high consumer (red area) categories on the basis of the normalized monthly bill per area using the k-means clustering algorithm.

### 3.2. Feature importance analysis

Machine Learning can be applied to solve complex problems efficiently (Said and Erradi, 2019), (Abdelhedi *et al.*, 2020), (Ayadi *et al.*, 2020),(Jabbar *et al.*, 2018). In this study, random forest (RF), a widely used machine learning technique (Wainberg, Alipanahi and Frey, 2016) is used. Accordingly, we used RF classifiers to conduct the feature importance analysis as they employ bootstrap aggregating (bagging) to sample the training dataset to establish a set of decision tree estimators (Breiman, 1996). The bagging method generates an ensemble of predictors considering random subsets of the main training dataset to enhance the generalization of the ensemble. We used three techniques to perform the feature importance analysis using the trained classifier.

**Gini impurity:**

The measure of the quality of split depends on metrics such as the Gini impurity, defined as:

$$G_i = 1 - \sum_{k=1}^{n} p_{i,k}^2 \quad (1)$$

Where $G_i$ represents the Gini impurity of node I, whereas $p_{i,k}$ represents the ratio of class k among all instances in node i. The cost function to train a decision tree on the basis of the Gini impurity and the CART algorithm to determine the optimum split is expressed by (Breiman, 1984):

$$J(k, t_k) = \frac{m_{left}}{m} G_{left} + \frac{m_{right}}{m} G_{right} \quad (2)$$

Where $J(k, t_k)$ represents the cost function based on feature k, threshold $t_k$, $G_{left/right}$ represents the impurity of the left or right subset, and $m_{left/right}$ denotes the number of instances on the left or right. Random sampling is applied to both the training dataset and the feature set in a random forest classifier. RF classifiers can measure the relative importance of features determined based on the extent to which a feature, on average, reduces the impurity measure based on Equation 4.

$$NI_i = w_i G_i - w_{right,i} G_{right,i} - w_{left,i} G_{left,i} \quad (3)$$

$$FI_i = \frac{\sum_{j: node\ j\ splits\ on\ feature\ i} NI_j}{\sum_{j \in all\ nodes} NI_j} \quad (4)$$

Where $NI_i$ denotes the importance of node i, w denotes the weighted number of instances, and $FI_i$ denotes the importance of feature I.

- Permutation importance measures an increase in the model's prediction error following the permutation of the values of the feature, consequently breaking the relationship between the feature and the true outcome. If shuffling its values increases the model error, the feature is considered important. In contrast, it is considered unimportant if the model error remains the same because it implies that the model made the prediction without using the feature.

- SHAP: The SHAP explanation method draws upon coalitional game theory to calculate Shapley values, which reflect how to fairly distribute the payout among the features (i.e., the prediction). A player can also be represented by a group of feature values. To illustrate, pixels can be grouped into superpixels to explain an image, and the prediction is distributed among them. The Shapley value explanation is shown as an additive feature attribution method.

We measured the relative importance of demographic, socioeconomic, and behavioral factors in relation to the consumption pattern using two random forest classifiers. We used Gini impurity, random patches method, and the bootstrap algorithm to train the random forest to train 500 decision tree classifiers, which comprised the random forest. The first model included five attributes as independent variables, namely monthly expenses, household income, ethnicity, gender, and age, to categorize the consumption category (high and low). Using a ten-fold cross-validation analysis, the random forest estimators categorize the two classes with an average accuracy of 93.5% (90-97.1%). Considering the random behavior of the feature importance procedure, we conducted 1,000 iterations of the analysis to obtain the average feature importance of the independent attributes. We applied the same procedure to the second independent variable set consisting of the perceived factors that improve self-responsibility, motives for adopting home energy efficiency, and consequence awareness. The average accuracy of the second classifier was 92.6% (90-95.1%) for the same cross validation.

**Error! Reference source not found.**, **Error! Reference source not found.**, and **Error! Reference source not found.** show the correlation between these three variables and the independent variables in the first random forest classifier in the following order of importance for the first (household expenses, age, ethnicity, gender, and income) and the second classifier (responsibility factor, energy efficiency motives, and awareness). Accordingly, household expenses, respondent age, and ethnicity group are sufficient for differentiating household energy consumption patterns. In contrast, consequence awareness is not crucial for energy consumption. Furthermore, the results indicated that household monthly expenses impact the energy consumption more than the income level. The household monthly expenses level consists of factors such as appliance use, appliance types, and building characteristics. Age and ethnicity group impact the energy consumption, but gender does not. **Error! Reference source not found.**, **Error! Reference source not found.**, and **Error! Reference source not found.** illustrate that consequence awareness does not impact aggregate household energy patterns.

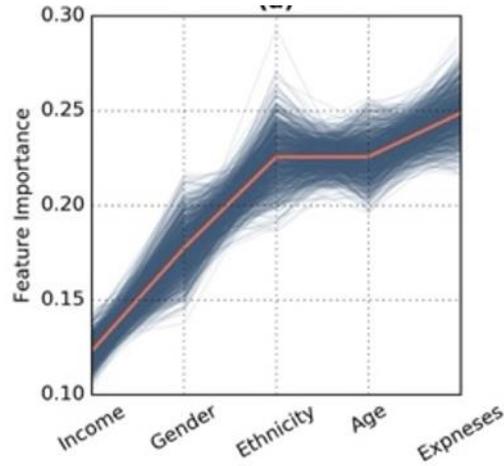

**Figure 4:** Feature importance analysis of the energy consumer categories based on demographic and socioeconomic factors using the Gini impurity.

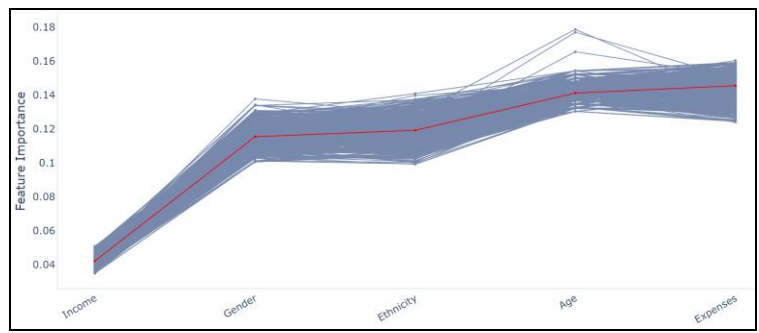

**Figure 5:** Feature importance analysis of the energy consumer categories based on demographic and socioeconomic factors using permutation importance.

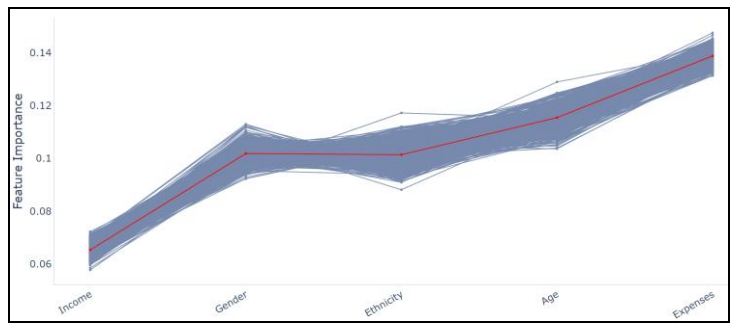

**Figure 6:** Feature importance analysis of the energy consumer categories based on demographic and socioeconomic factors using SHAP.

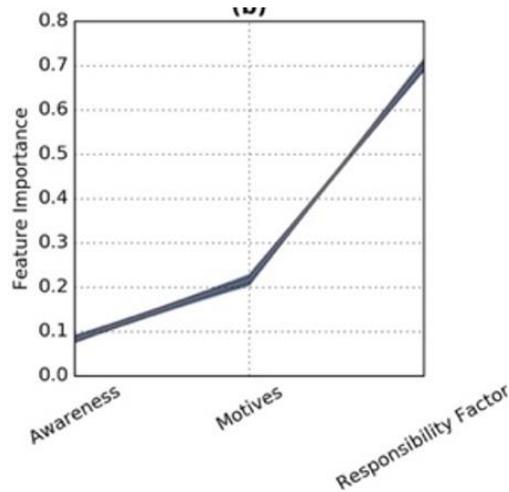

**Figure 7:** Feature importance analysis of the energy consumer categories based on human attitudes and behavioral factors using the Gini impurity.

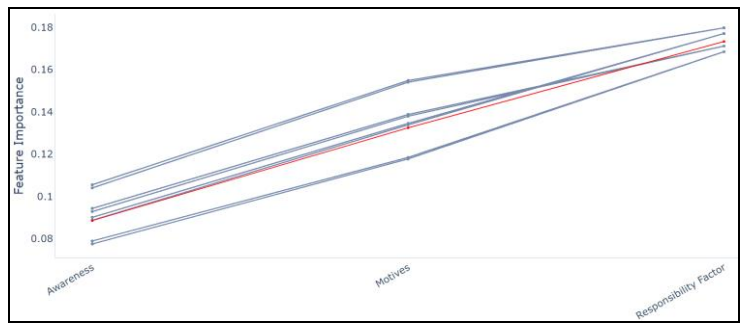

**Figure 8:** Feature importance analysis of the energy consumer categories based on human attitudes and behavioral factors using permutation importance.

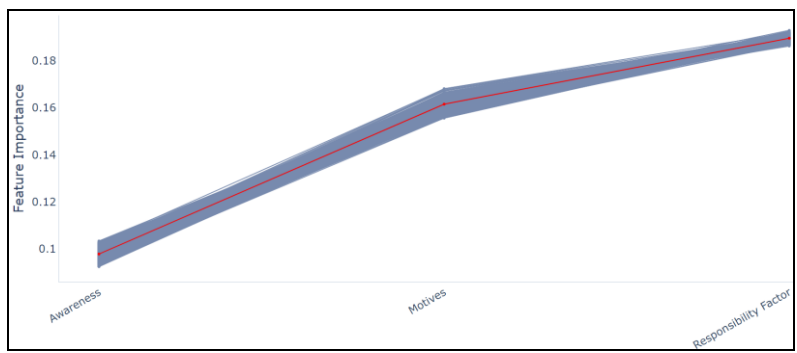

**Figure 9:** Feature importance analysis of the energy consumer categories based on human attitude and behavioral factors using SHAP.

Figure 10 shows the associations between the investigated variables and energy consumption categories. Arabic and North American groups and older people tend to have high consumption patterns, in addition to those motivated by financial incentives and those concerned about home energy efficiency.

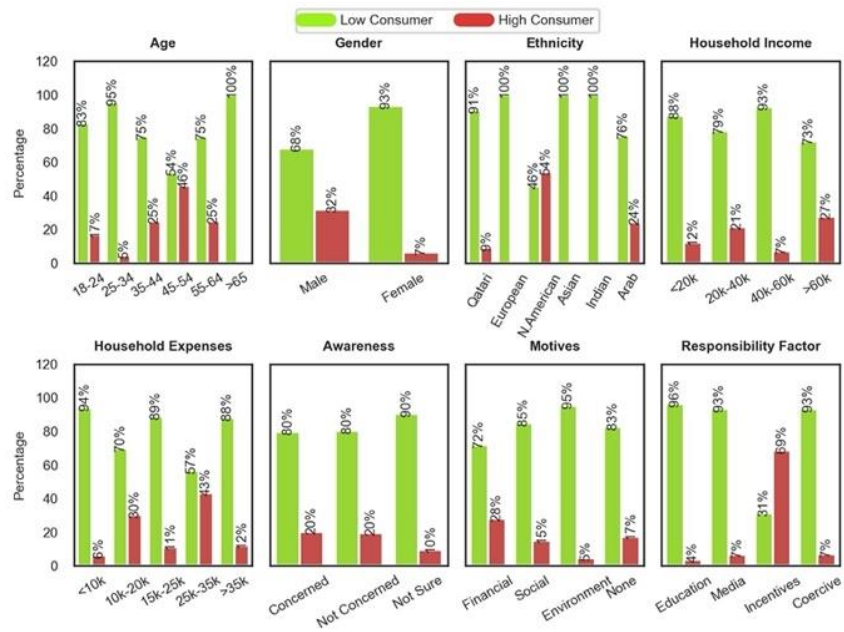

Figure 10: Energy consumer categories demonstrated based on human-oriented factors.

## 4. Discussion

We investigated the associations between the factors regarding: (1) thermal comfort in residential buildings and human-building interactions, (2) productivity and human-building interactions in workplaces, (3) behavioral factors and energy attitudes, and (4) financial drivers in residential buildings.

### 4.1. Analysis of thermal comfort in residential buildings

Heating, ventilation, and air conditioning (HVAC) systems keep the indoor environment comfortable, and they account for a considerable amount of energy consumption in hot climates as they largely depend on user's preferences. The saving energy in some cases is in contrast with the user's well-being. Studies have demonstrated that the well-being and preferences regarding the indoor environment are in correlation with socioeconomic and demographic factors (e.g., (Schweiker et al., 2018; Sintov, White and Walpole, 2019)). Violin plots based on KDE distributions of the respondents' perceptions about thermal comfort at home and the importance of the comfort level for their well-being are shown in **Error! Reference source not found.**. The right column illustrates the importance of indoor thermal comfort for their well-being based on ethnicity, income, and gender. Users' perceptions of comfort level are presented in the left column. Comfort levels of women tend to be higher, implying that they feel more comfortable at home than men. Moreover, it can be seen from the right column that women value more indoor comfort than men. Considering the household income level, it is in correlation with comfort importance and perceived comfort. Furthermore, Qataris, Arabs, and Europeans value indoor comfort more than North American, Asian, and Indian ethnic groups. Accordingly, strategies in electricity pricing and financial incentive modeling can be proposed to enhance the well-being of communities. Quantifying penalty factors for comfort deviation and building design, operation, and control guidelines is also possible. Correlations between the occupant's comfort level and building construction type and conditions were determined as well. **Error! Reference source not found.** shows a general enhancement of the indoor comfort level for newer buildings. Moreover, users are in general more satisfied in Arabic-style homes and villas than in the flats. In addition, modern buildings offer more comfortable indoor environments. Finally, according to the comparison results, central ducted systems and wall-mounted units provide higher indoor comfort compared with window air conditioners and ceiling fans.

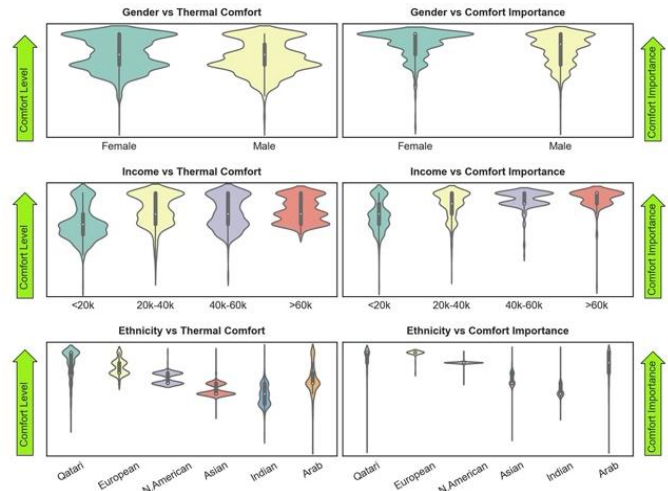

**Figure 11: Associations between demographic factors and participants' perceived thermal comfort at home (left column plots) and indoor comfort importance for the respondents' quality of life (right column plots) based on Kernel Density Estimation.**

Accordingly, energy efficiency and retrofit financial incentives can be directed toward systems providing the lowest comfort levels (and least efficiency) to enhance the overall well-being of society and energy efficiency indicators.

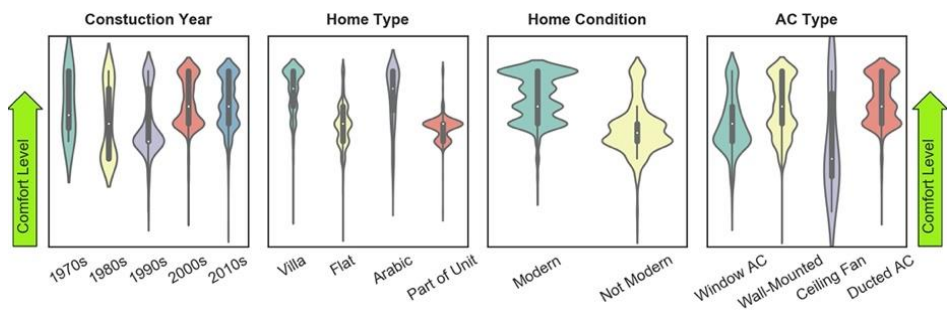

**Figure 12: Correlation between perceived thermal comfort at home and dwelling characteristics.**

## 4.2. Analysis of thermal comfort in workplaces

In addition, it is possible to use visual and thermal comfort measures to determine lower consumption modes and at the same time save energy while meeting user's preferences, According to **Error! Reference source not found.**, ethnicity and gender impact perceived comfort in the indoor environment in workplaces. Men prefer higher temperatures (22-24 °C) than women (below 22 °C). Considering ethnicities, Asians prefer hot indoor environments compared with North Americans and Arabs. Asians are inclined to change thermostat settings and open windows, in contrast to women. Also, Asian ethnic groups reported preferences toward light workplaces. Patterns in human-building interactions are shown in **Error! Reference source not found.**.

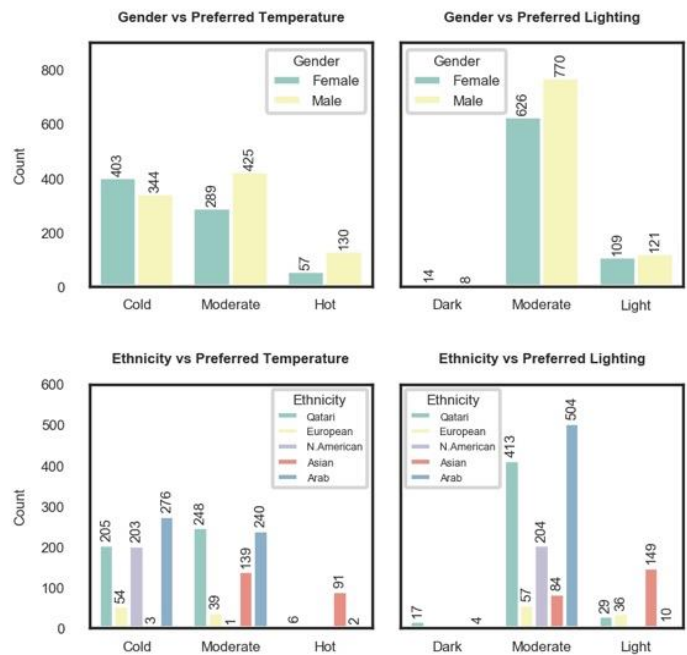

**Figure 13: Intercorrelations between the demographic factors and users' preferences in the workplace indoor environment.**

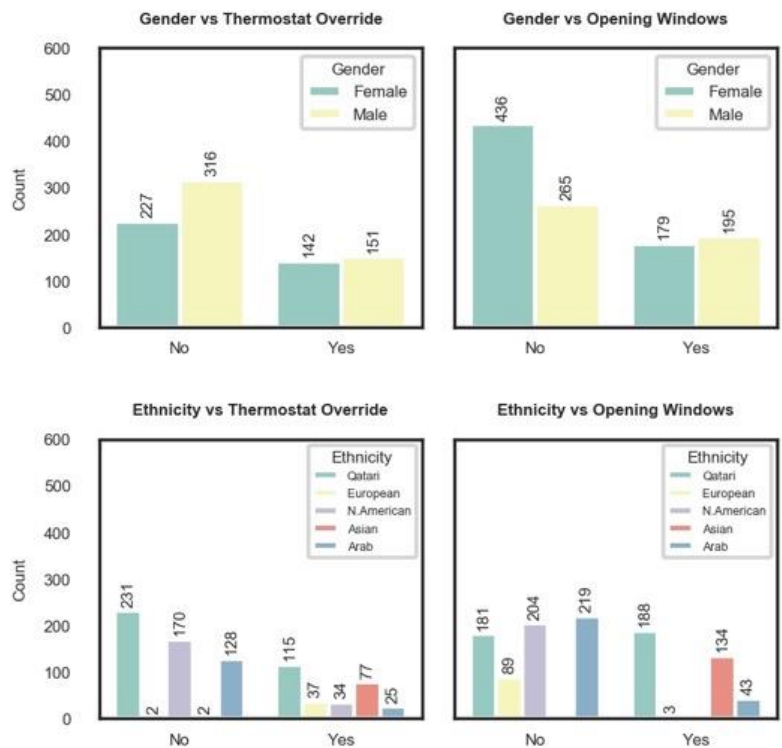

**Figure 14: Intercorrelations between the demographic factors and users' interactions with the building.**

**Error! Reference source not found.** shows patterns concerning the respondents' perceived comfort at work and the impact of the comfort level on productivity and performance. Women value thermal comfort more than men, which is in line with the findings relayed to the perceived comfort at home. The results also indicated a positive correlation between income and impact on productivity and perceived comfort at work. Regarding ethnicity, Asians and Indians reported less comfort, although it is a critical factor for productivity. In addition, the correlations between human-building interactions and indoor environment preference with perceived comfort level at work are shown in **Error! Reference source not found.**. According to the results, individuals with hotter indoor environment preferences reported more discomfort.

The results also revealed that those who frequently open windows and override the thermostat tend to feel higher thermal discomfort, which implies the importance of thermal discomfort in preventing negative human interruptions. The influence of building construction and attributes on the perceived comfort level is shown in **Error! Reference source not found.**. Notably, users of personal offices have a higher perceived comfort level at work than those in open office spaces.

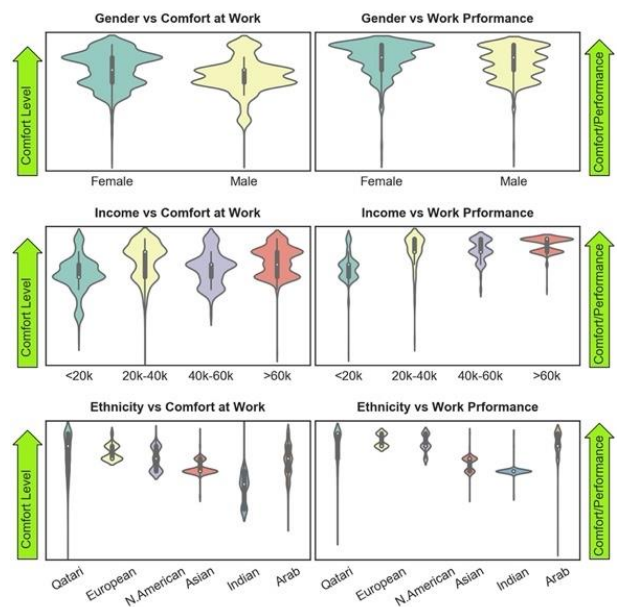

**Figure 15: Associations between demographic factors and participants' perceived thermal comfort at work (left column plots) and indoor comfort importance on the respondent's work performance (right column plots).**

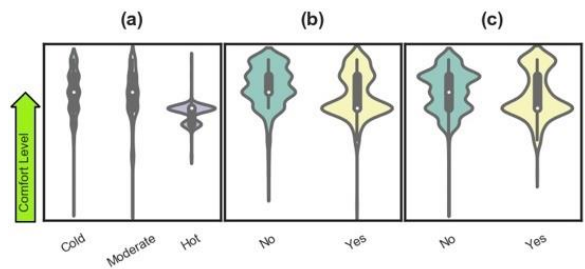

**Figure 16: Relationship between the respondents' perceived comfort with (a) a preferred indoor temperature, (b) tendency toward thermostat override, and (c) tendency toward opening windows.**

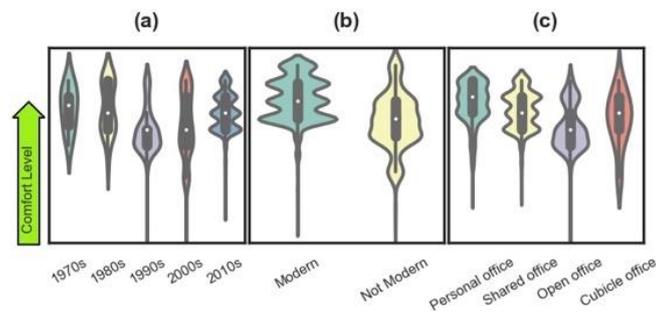

**Figure 17: Relationship between the respondent's perceived comfort and (a) workplace construction year, (b) workplace style, and (c) workplace type.**

### 4.3. Awareness, motives, and responsibility

We also assessed the associations between human dimensions and factors in terms of perceived energy consumption, perceived factors that can enhance energy consumption responsibility, motives for home energy efficiency, and climate change awareness. This aimed at assisting decision-making in proposing strategies for different population segments, such as developing personal and social norms and raising awareness. The associations between ethnicity, income, and gender with consequence awareness are shown in the first-row plots of **Error! Reference source not found.**. The highest income group and women are the most concerned about climate change, in contrast to Indian and Arab ethnic groups. Men value more financial motives for home energy efficiency, as shown in the second-row plots of **Error! Reference source not found.**. Moreover, this figure also shows that the lower-income groups reported financial factors, Qataris social and environmental factors, and other ethnic groups the financial factors as the main driver in relation to home energy efficiency. The coercive factors were more prominent among male respondents (the third-row plots). North Americans highly value financial incentives, Europeans value education, and other ethnic groups coercive actions (Nazemi, Jafari and Zaidan, 2021). Female respondents reported considerably more responsible behavior regarding self-perceived energy responsibility at work (fourth-row plots), especially in comparison with Indian and Asian ethnic groups. The highest proportion of adverse interactions in work environments was reported in the lowest-income group.

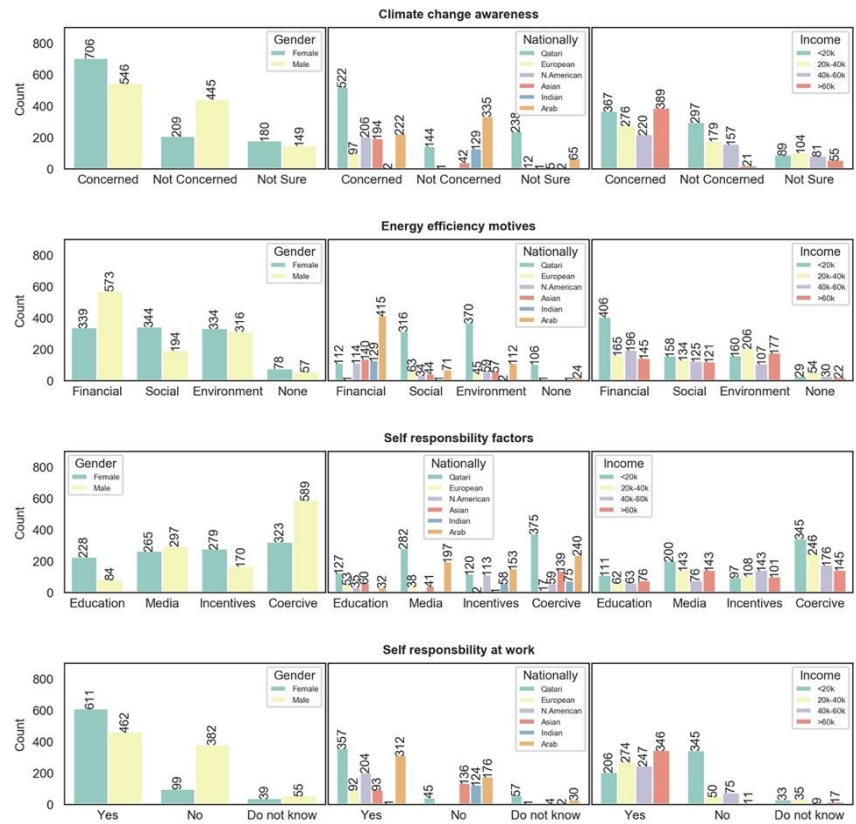

**Figure 18: Intercorrelation between respondents' attitudes and key demographic/socioeconomic factors.**

### 4.4. Financial drivers in energy consumption patterns

Determining the primary financial drivers of home energy consumption patterns and their associations with human factors allow developing energy strategies. To illustrate, determining if high electricity prices of energy subsidies modify energy consumption behavior can inform such strategies. The impact of income, ethnicity, and gender on energy pricing and energy use associations is shown in **Error! Reference source not found.**. Thus, high electricity prices can considerably impact men to modify their energy consumption, in particular those from Indian, Asian, and Arab ethnic groups. Moreover, electricity pricing is more effective strategy for reducing consumption among low-income than high-income respondents. In the second row of **Error! Reference source not found.**, the association between human factors and willingness to participate in DR programs is illustrated. For women, social and environmental incentives are more important for participating in DR programs, whereas men emphasize financial incentives. Regarding ethnicities, Qataris, in general, are against participation in DR programs, while other ethnic groups reported financial incentives as the most important factor. An association between DR participation and the household income level was not determined. The associations between the income level, ethnicity, and gender and energy consumption as a result of waived electricity bills are shown in the last row. Considering gender, women reported that they would not change their energy use due to waived electricity bills. Indian and Arab ethnic groups reported that they would increase the usage in the case of waived electricity bills, in contrast to other ethnic groups. Finally, lower-income groups would likely increase energy consumption, in contrast to high-income respondents.

Findings regarding ratepayers' willingness to participate in demand response programs are shown in **Error! Reference source not found.**. In contrast to villa owners, apartment residents are much more inclined to participate in DR. Moreover, there is a distinction between homeowners who are not interested in DR participation and home rental residents who are. Finally, the residents of old-style buildings are more inclined toward DR participation.

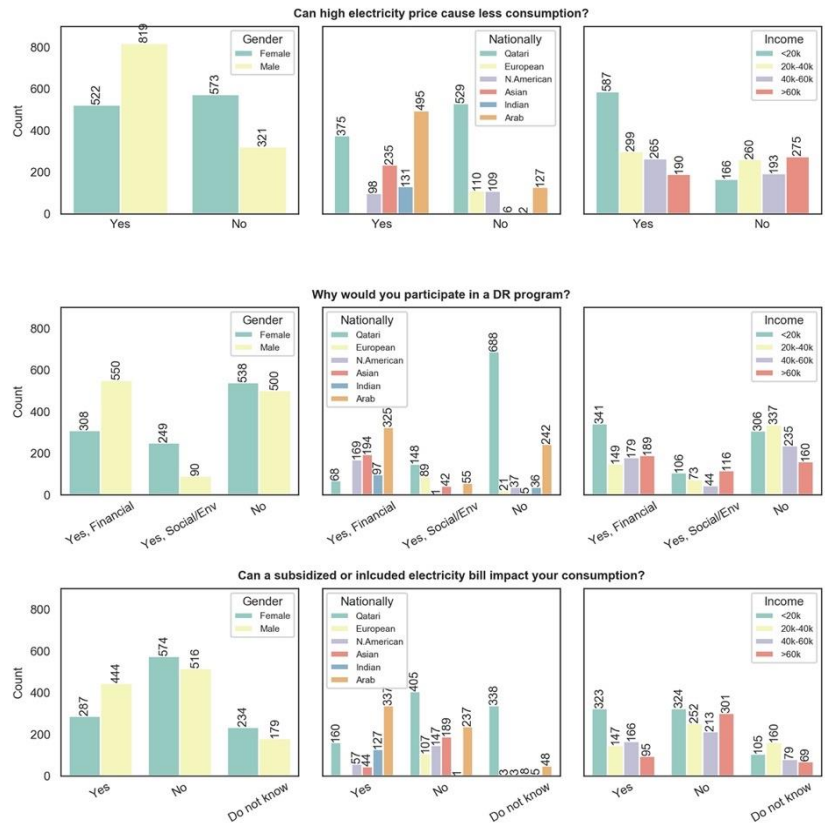

**Figure 19:** Interdependencies between demographic/socioeconomic factors and financial drivers of home energy consumption and DR participation willingness.

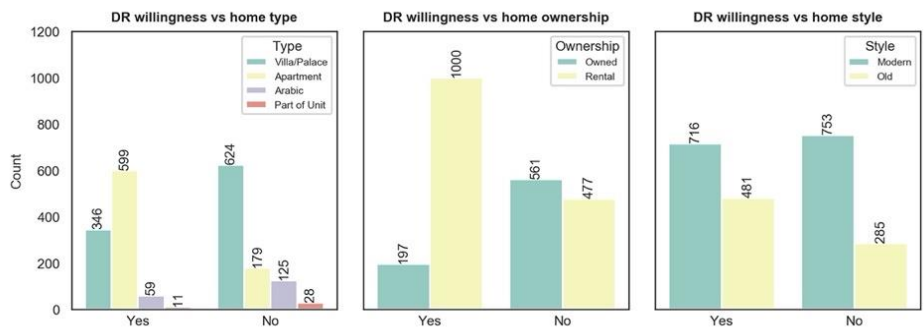

**Figure 20:** DR willingness association with building attributes with respect to home type, ownership, and style.

## 5. Conclusion

In this study, we carried out an empirical and analytical study by surveying a sample of 2,200 respondents in Doha, Qatar with an aim to investigate the influence of behavioral, socioeconomic, and demographic factors on human-building interactions. We employed the machine learning method to categorize the respondents into high and low ratepayer classes on the basis of normalized energy use per area. The feature relevance analysis revealed that the most important factors needed to determine family consumption patterns are ethnicity, age structure, and household costs. In addition, the results indicated that behavioral traits and human attitudes can be used to categorize the consumption patterns. Owing to these findings, the decision-makers can develop strategies aimed at raising awareness among particular social communities. Moreover, these findings can contribute to proposing interventions aimed at reducing high consumption among specific groups. Furthermore, the findings indicated the considerable association between personal well-being and comfort levels in the indoor environment. Importantly, we also investigated how socio-economic and demographic factors and building characteristics impact human-building interactions. Therefore, one important implication of this study is that it is necessary to define incentives (e.g., financial) aimed at increasing well-being in the indoor environment and minimizing adverse human-building interactions by investigating and determining specific community characteristics. Finally, the results indicated that factors such as income level, ethnicity, and gender impact users' willingness to participate in demand response programs with a monetary incentive, as well as the financial drivers regarding residential energy consumption. Owners of older-style buildings, apartment occupants, and home rental residents reported higher inclination to participate in DR projects, in contrast to homeowners who strongly opposed them.

This study has some limitations. First, it did not take into account the COVID-19 pandemic and its possible impacts on electricity consumption trends. Therefore, we recommend conducting studies that will take into account new data updates resulting from the pandemic. Furthermore, this study divided the respondents into only two classes: high and low consumers. Therefore, we suggest including a more detailed decomposition analysis to fully capture consumption trends of different end-user groups and determine related factors.

## 6. Acknowledgments

This publication was made possible by an NPRP award [NPRP11S-1228-170142] from the Qatar National Research Fund (a member of Qatar Foundation). The statements made herein are solely the responsibility of the authors. The publication of this article was funded by the Qatar National Library